\documentclass[
]{ceurart}

\usepackage{listings}
\usepackage{inputenc}
\usepackage{makecell} 
\usepackage{subcaption}
\usepackage{xurl} 
\usepackage{hyperref}
\begin{document}

\copyrightyear{2026}
\copyrightclause{Copyright for this paper by its authors.
  Use permitted under Creative Commons License Attribution 4.0
  International (CC BY 4.0).}

\conference{MediaEval'26: Multimedia Evaluation Workshop,
  June 15-16, co-located with ACM ICMR 2026, Amsterdam, The Netherlands}

\title{A Test-time Actor-Critic Approach to News Images Generation}


\author[1]{Damianos Galanopoulos}[%
    email=dgalanop@iti.gr,
]

\author[1]{Vasileios Mezaris}[%
    email=bmezaris@iti.gr,
]

\address[1]{Information Technologies Institute (ITI),
Centre of Research and Technology Hellas (CERTH),
Thessaloniki, Greece}

\begin{abstract}
 This paper introduces the CERTH-ITI solution for the NewsImages 2026 challenge, which focuses on generating images related to news headlines. Inspired by the Actor-Critic paradigm in reinforcement learning, we present a test-time, model-agnostic Actor-Critic Image Generation approach (ACIG). ACIG generates prompts for image creation, produces the images, evaluates the generated results, and if needed refines the image generation prompts accordingly in a feedback loop.
\end{abstract}
\maketitle

\section{Introduction}\label{sec:intro}
The MediaEval NewsImages 2026 challenge~\cite{heitz2026newsimages, heitz2025newsimages, heitz2024empirical} focuses on generating and/or retrieving images that appropriately illustrate a news item, based only on the news item's headline. Contrary to our previous efforts~\cite{Galanopoulos2025cross, leventakis2023cross}, which were in the direction of image retrieval, this year we tackle the image generation task. For NewsImages 2026, we, the CERTH-ITI team\footnote{The code is available at \url{https://github.com/IDT-ITI/actor-critic-image-generation}.} propose Actor-Critic Image Generation (ACIG), a training-free pipeline for generating images from article headlines. Inspired by the Actor-Critic paradigm in reinforcement learning, ACIG iteratively produces, evaluates, and refines image generation prompts through a closed feedback loop, requiring no gradient updates or fine-tuning of any component model.

\section{Related Work}\label{sec:work}

Our ACIG approach leverages test-time optimization to address the challenges of the MediaEval 2026 NewsImages task, particularly the alignment between visual outputs and news headlines. 
The paradigm of agentic methods, in which models autonomously iterate and complete tasks through self-correction, has gained significant momentum. Recent research \cite{Jang_2026_CVPR, GenPilot25} demonstrates that Vision-Language Models (VLMs) can serve as effective ``critics'' to refine text-to-image alignment in real time by analyzing visual-textual correspondence. In contrast to previous approaches that rely on heavy reinforcement learning (RL) training cycles, the ACIG pipeline adopts a test-time optimization strategy. Moreover, effective prompt engineering is increasingly viewed as a creative, iterative process rather than a single instruction \cite{Oppenlaender18082025}. This iterative philosophy is further supported by the PACE framework \cite{dong2024pace}, which demonstrates the effectiveness of Actor-Critic editing in refining prompts for large language models and ensuring that generated outputs remain faithful to intended semantic constraints. This modular, agent-led workflow represents a significant advancement in bridging the gap between abstract headline narratives and the concrete visual representations required in modern newsrooms \cite{GenPilot25}.

\section{Approach}\label{sec:approach}

\subsection{Pipeline Overview}

Given a set of news article headlines $\mathcal{A} = \{a_1, a_2, \ldots, a_N\}$, ACIG processes each article independently over up to $T$ iterations. At each iteration $t$, the pipeline goes through three sequential stages: prompt generation (Actor), image synthesis (Image Generator), and quality assessment (Critic). In the latter stage, if an image's critic score meets or exceeds a quality threshold $Q$, the generation process for the corresponding article is marked as complete, and no further iterations are performed. If $T$ iterations are completed without meeting quality threshold $Q$, the final output is the generated image that received the highest critic score across all iterations.

\subsection{Actor: Prompt Generation}

The Actor is a large vision-language model (VLM), i.e., Qwen3-VL-8B-Instruct \cite{qwen3technicalreport, Qwen2.5-VL}, responsible for transforming a headline into a descriptive prompt suitable for image-generation models. At iteration $t=0$, the Actor receives the article headline $a_i$ as context and is prompted to produce a single image generation prompt $p_i^{(0)}$, with an explicit instruction to emphasise a non-photorealistic visual style. At subsequent iterations $t > 0$, the Actor is additionally provided with the full scoring history from all prior attempts:
 \begin{equation}
    \mathcal{H}_i^{(t)} = \left\{ \left(p_i^{(j)},\ \left\{ s_i^{(j,k)} \right\}_{k=1}^{K} \right) \right\}_{j=0}^{t-1}
\end{equation}

\noindent where $s_i^{(j,k)}$ denotes the critic score assigned to the image produced by the $k$-th generation model at attempt $j$, and $K$ is the total number of image generators active in the run. This history is formatted as a structured string and passed to the Actor, which is instructed to generate an improved prompt with awareness of previous failures. 

\subsection{Image Generator: Candidate Images Synthesis}

Each Actor-generated prompt is forwarded to one or more diffusion-based image generation models. ACIG is designed to be model-agnostic at this stage; in our experiments, we evaluate three generators: Z-Image-Turbo \cite{jiang2025distribution, team2025zimage}, Qwen-Image (SDNQ uint4 quantised), and Qwen-Image-2512 \cite{qwenimage2025}. For brevity, in Table~\ref{tab:official} we refer to these models as ZT, SDNQ, and 2512, respectively. All generators produce images in a 16:9 aspect ratio, and a shared base seed is used across runs. As all of these generators support negative prompting, a fixed negative prompt is also supplied to suppress common artifacts such as low resolution, anatomical distortions, and AI-characteristic over-smoothness.
 
\subsection{Critic: Image Quality Assessment}

The Critic is a separate VLM, Qwen3.5-9B \cite{qwen3.5}, tasked with evaluating the relevance of each generated image with respect to its source headline. For each image in the current iteration $t$, the Critic is queried to rate the generated images on a Likert [1-5] scale on how accurately the image captures the article's headline. The model is constrained to return a single integer. The resulting scores $\{s_i^{(t,k)}\}$ - one per image per article - are stored in the corresponding prompt entry, and the maximum score across all images of a given attempt,

\begin{equation}
    s_i^{(t)} = \max_{k}\ s_i^{(t,k)},
\end{equation}

\noindent is used to determine if the iterative process should be concluded ($s_i^{(t)}\geq Q$) or continued.

\subsection{Duplicates}
In order to fully comply with the task rules, after the final image per article headline is generated, we compute the MD5 hashes of all image and identify duplicate images for the same article across runs. For these duplicates, starting from run \#10, we re-run the procedure for the affected articles by altering the seed on the generation models. This process continues sequentially through the remaining runs down to run \#2, ensuring that all submitted images are globally unique across the full set of articles.

\section{Submitted Runs and Results}
\label{sec:results}

\begin{table}[!t]
\centering
\caption{Configuration of submitted runs; Results (Average ratings) from the official collaborative online evaluation event and our internal experiments. The best /
second-best results are in bold / underline; higher rating values are better; and, per-run computational efficiency statistics are also reported. Column $\mathcal{H}^{(t)}$ indicates whether the critic scoring history is passed to the Actor for prompt refinement; in the contrastive runs where this is disabled (``---''), the Actor is asked to re-generate a prompt from the headline alone at each iteration. $T=0$  denotes one-shot generation, without any Actor-Critic iterations.}
\label{tab:official}
\resizebox{\textwidth}{!}{\begin{tabular}{lcccc|cc|ccc}
\toprule
\multirow{3}{*}{\textbf{Run}} & \multirow{3}{*}{\makecell{\textbf{Image} \\ \textbf{Generator(s)}}} & \multirow{3}{*}{\makecell{\textbf{Qual. } \\ \textbf{Thres.} \\ $Q$}} & \multirow{3}{*}{\makecell{\textbf{Iter.} \\ $T$}} & \multirow{3}{*}{\makecell{\textbf{Hist.}\\ $\mathcal{H}^{(t)}$}}   & \multicolumn{2}{c|}{\textbf{AVG rating $\uparrow$ }} &  \multirow{3}{*}{\makecell{\textbf{\# of } \\ \textbf{ iter.}\\ \textbf{/article}}$\downarrow$} & 
\multirow{3}{*}{\makecell{\textbf{Time}\\ \textbf{(sec.)} \\ \textbf{/iter.}}$\downarrow$} & 
\multirow{3}{*}{\makecell{\textbf{Total time}  \\\textbf{(sec.)} \\ \textbf{/article}}$\downarrow$} \\

 \cmidrule(lr){6-7}
  & & & & & \multirow{2}{*}{\makecell{\textbf{Int.}}} & \multirow{2}{*}{\makecell{\textbf{Offic.}}}  & & & \\
   & & & & &  &  & & & \\
   
\midrule

\#1  & ZT/2512/SDNQ & 4 & 5 & \checkmark    &\textbf{4.86}      &  3.429                & 1.22   & 91.8 + 3.2  & 115.9 \\
\#2  & ZT/2512/SDNQ & 5 & 5 & \checkmark    & \underline{4.84}  & \underline{3.517}     & 3.60   & 91.8 + 3.2  & 342 \\
\#3  & ZT/2512/SDNQ & 5 & 5 & ---           & 4.74              &  3.362                & 3.57   & 91.8 + 0.9  & 330.9\\
\#4  & ZT/2512/SDNQ & ---  & 0 & ---        & 4.68              &  3.431                & 1     & 91.8  & 91.8\\
\midrule
    \#5  & ZT           & 4 & 5 & \checkmark    & 4.72              & 3.379             & 1.64   & 1.2 + 1.9   & 5.1 \\
\#6  & ZT           & 5 & 5 & \checkmark    & 4.80              &  \textbf{3.616}       & 4.60   & 1.2 + 1.9   & 14.3 \\
\#7  & ZT           & 5 & 5 & ---           & 4.70              &   3.300               & 4.54   & 1.2 + 0.3   & 6.8\\
\midrule
\#8  & SDNQ        & --- & 0 & ---          & 4.53              &  3.197                & 1     & 55.3  & 55.3 \\
\#9  & 2512        & --- & 0 & ---          & 4.60              &   3.355               & 1     & 35.3  & 35.3 \\
\#10 & ZT          & --- & 0 & ---          & 4.51              &   3.350               & 1     & 1.2   & 1.2\\
\bottomrule
\end{tabular}}
\end{table}

In Table~\ref{tab:official} we present the configuration of the ten submitted runs and whether prompt history and scoring were used during generation. We also report the results from both the official NewsImages 2026 evaluation, and an internal evaluation that we ran. For the latter, we utilized the NewsImages 2025 SMALL test dataset \cite{heitz2025newsimages}, and we assessed in-house each generated image similarly to the official evaluation protocol, on a Likert [1-5] scale. Moreover, to assess computational efficiency, we report: the average number of iterations per input article, the average time per iteration (in sec., for image generator + Actor-Critic; the latency of the Actor when $\mathcal{H}^{(t)}$ is disabled is negligible), and the total generation time per input article on a Nvidia RTX4090 GPU.

The results show that enabling critic scoring history $\mathcal{H}^{(t)}$ consistently improves performance across both evaluation settings. 
The two best-performing runs overall (\#2 and \#6) both use $Q=5$ with history enabled. Disabling critic scoring history yields a noticeable performance drop in both internal and official ratings. The non-ACIG single-shot baselines (\#4, \#8, \#9, \#10) are generally outperformed by the iterative runs with active feedback, though \#4 remains relatively competitive on the official metric. 
Overall, using a single well-performing image generation model (ZT) within the proposed ACIG approach (run \#6) yields the best results, both among our runs and among all NewsImages 2026 challenge submissions, according to the leaderboard, while also achieving very competitive computational efficiency.

\begin{figure}
\centering
\begin{subfigure}{0.48\textwidth}
    \includegraphics[width=\textwidth]{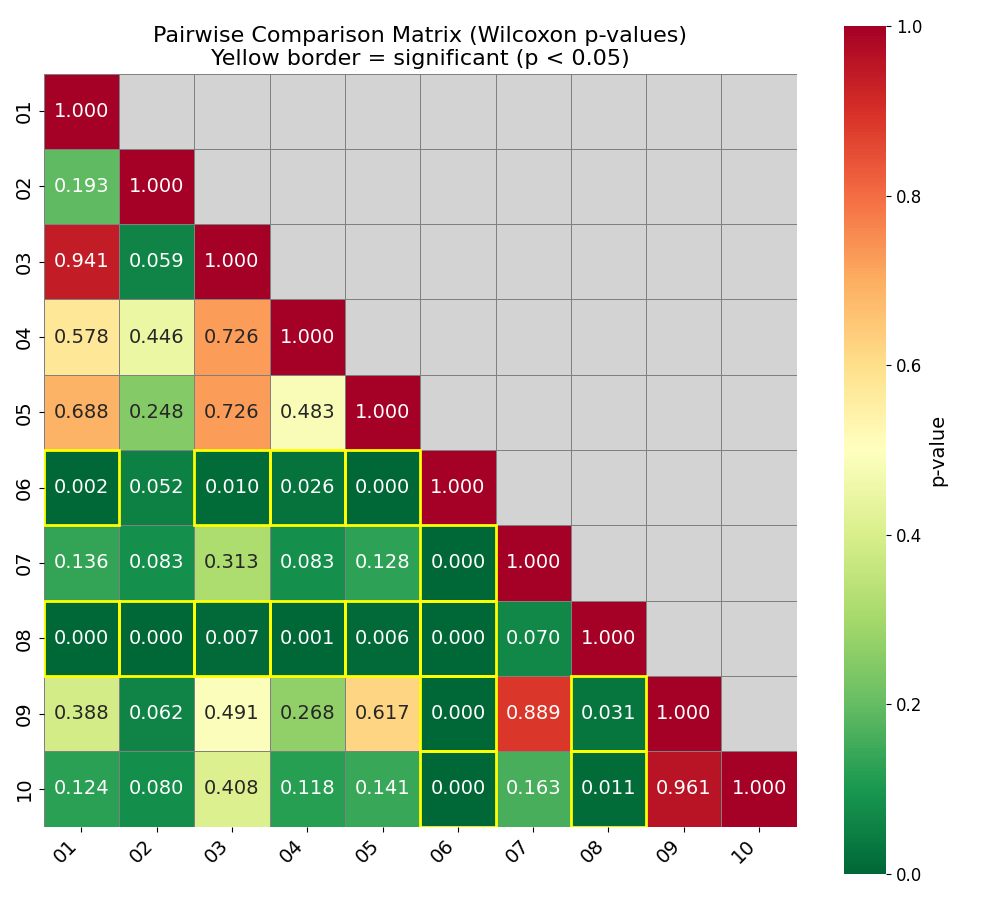}
    \caption{Official results}
\end{subfigure}
\hfill
\begin{subfigure}{0.48\textwidth}
    \includegraphics[width=\textwidth]{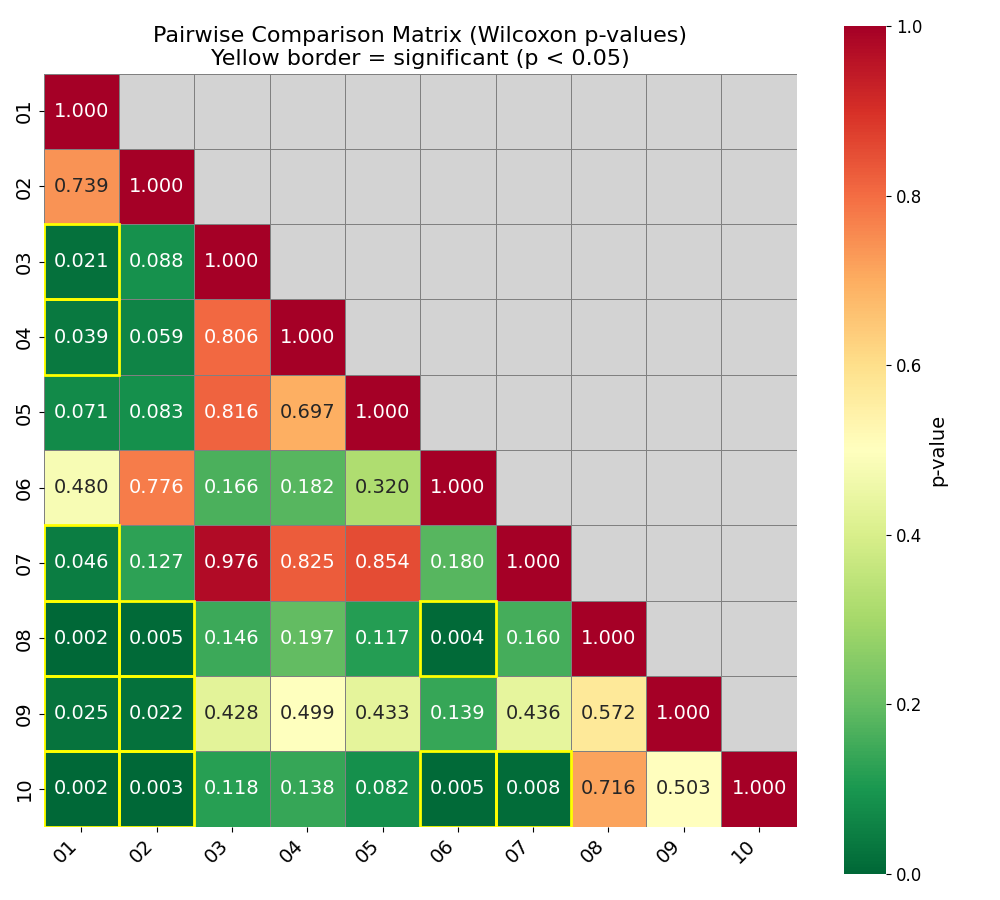}
    \caption{Internal results}
\end{subfigure}

\caption{Pairwise Wilcoxon signed-rank test $p$-values, for all pairwise comparisons between the 10 evaluated runs. Yellow borders in a cell indicate a pair of runs whose performance difference is statistically significant ($p< 0.05$).}
  \label{fig:ptest}
\end{figure}

In Figure \ref{fig:ptest}, we present the pairwise Wilcoxon signed-rank test p-values~\cite{wilcoxon1945individual}, comparing human evaluation scores across runs. For each pair of runs $(x, y)$, let $d_i^{(x,y)} = r_i^{(x)} - r_i^{(y)}$ denote the difference in ratings for article $i$. The test statistic is $W^{(x,y)} = \sum_{i=1}^{N} (\text{sgn}(d_i^{(x,y)}) \cdot R_i^{(x,y)}) $, where $R_i^{(x,y)}$ is the rank of $|d_i^{(x,y)}|$ among all non-zero differences (in ascending order), and $\text{sgn}(\cdot)$ is the sign function. The p-value is derived from the distribution of $W^{(x,y)}$ under the null hypothesis of no difference; $p^{(x,y)}<0.05$ indicates a statistically significant difference. Although most p-values in Fig. 1 exceed $0.05$, due to the small number of evaluated images, our core finding that ACIG runs \#6 and \#2 outperform the non-ACIG single-shot baselines (runs \#8, \#9, \#10) is supported by, in most cases, statistically significant differences.

\section{Conclusion}
In this paper we presented a test-time, model-agnostic Actor-Critic Image Generation approach. We showed the effectiveness of iterative refinement over single-pass generation, demonstrating that the proposed ACIG Actor-Critic approach yields images that match considerably better the provided textual prompt (article headline). Within ACIG, a single lightweight image generator (ZT) can exceed the performance of much more computationally expensive alternatives.

\section*{Acknowledgements}
This work was supported by the EU’s Horizon Europe programme under grant agreement 101214398 ELLIOT.
\clearpage
\section*{Declaration on Generative AI}
 
 During the preparation of this work, the authors used Grammarly to Perform Grammar and spelling checks. After using this tool, the authors reviewed and edited the content as needed and take full responsibility for the publication’s content.

\def\bibfont{\small} 

\clearpage

\appendix
\section{Appendix}

\subsection{Image Generation Prompts}
\subsubsection*{Initial Prompt ($t=0$) }
\begin{quote}
``Using this article title `$article\_title$' create a prompt for an image generation model to create a relevant to the article image. Return ONLY the prompts. Stress that the image should be non-photorealistic.''
\end{quote}

\subsubsection*{Negative Prompts}
\begin{quote}
``Low resolution, low image quality, limb deformities, finger deformities, oversaturated image, wax figure appearance, lack of facial detail, excessive smoothing, AI-like appearance. Chaotic composition. Blurry and distorted text.''
\end{quote}

\subsubsection*{Refinement Prompt}\begin{quote}
``You are a prompt optimization expert. Your goal is to generate an image generation prompt that best captures the news article with title `$article\_title$'. Here is the history of previous attempts and their scores (1–5, higher is better): `$history\_str$'. Analyze what made higher-scoring prompts better and lower-scoring ones worse. Generate a new prompt that improves on these attempts. Output only the prompt, with no explanation or additional text.''
\end{quote}

 \subsubsection*{Example of History $\mathcal{H}^{(t)}$}

An example of the history $\mathcal{H}^{(t)}$, denoted as `$history\_str$' in the refinement prompt provided in the previous paragraph, for $t=2$, is given below. The original article headline $a_i$ that this example corresponds to is: ``US Dollar Outlook: GBP/USD May Fall as USD/CAD Rises Amid Changes in Retail Exposure''.
 
\begin{quote}
``Attempt 1: Prompt: \textit{`Non-photorealistic illustration of a financial chart with fluctuating lines representing GBP/USD and USD/CAD exchange rates, surrounded by stylized currency symbols, digital circuit patterns, and abstract market indicators, conveying dynamic currency movement and retail exposure shifts — vibrant, modern, and symbolic, not photorealistic.'} | Score: 2 

Attempt 2: Prompt: \textit{`Dynamic financial dashboard visualization showing GBP/USD trending downward and USD/CAD trending upward, with glowing data streams and interactive chart elements, surrounded by stylized retail investors and currency icons, conveying real-time market shifts due to retail exposure changes — sleek, high-tech, data-driven, and visually engaging, optimized for clarity and economic storytelling.' }| Score: 3''
\end{quote}

\subsection{Image Quality Assessment}
\begin{quote}
``Rate 1–5 how accurately this image captures the key attributes of the text article without depicting any important elements not present in the article given the headline: `$article\_title$'. Reply with a single integer between 1 and 5. Do not explain. Do not add any other text. Your entire response must be exactly one digit. Provide your rate using a single score.''
\end{quote}

\clearpage
\subsection{Model URLs}

\begin{table}[h]
\centering
\caption{LLMs and image generation models used in ACIG}
\begin{tabular}{ll}
\toprule
\textbf{Model} & \textbf{URL / HuggingFace ID} \\
\midrule
Qwen3-VL-8B-Instruct  & \url{https://huggingface.co/Qwen/Qwen3-VL-8B-Instruct} \\
Qwen3.5-9B            & \url{https://huggingface.co/Qwen/Qwen3.5-9B} \\
Z-Image-Turbo         & \url{https://huggingface.co/Tongyi-MAI/Z-Image-Turbo} \\
\multirow{2}{*}{\makecell[l]{Qwen-Image \\ (SDNQ uint4 quantised)}} & 
\multirow{2}{*}{\makecell[l]{\url{https://huggingface.co/Disty0/Qwen-Image-SDNQ-uint4-svd-r32}}} \\
& \\
Qwen-Image-2512       & \url{https://huggingface.co/Qwen/Qwen-Image-2512} \\
\bottomrule
\end{tabular}
\end{table}

\subsection{Qualitative Results}

\begin{table*}[ht]
\centering
\caption{Iterative Actor-Critic image generation for the headline \textit{``Surfers Share Waves With Blue Whale in Dana Point.''}}
\label{tab:iterations_16498}
\setlength{\tabcolsep}{4pt}
\renewcommand{\arraystretch}{1.3}
\begin{tabular}{m{0.3cm} m{8.5cm} m{3.5cm} c  }
\toprule
\textbf{$t$} & \multicolumn{1}{>{\centering\arraybackslash}m{8.5cm}}{\textbf{Actor-produced prompt}} & {\makecell{\textbf{Generated} \\ \textbf{image}}}   & {\makecell{\textbf{Critic's} \\ \textbf{score}}} \\
\midrule

0 & \footnotesize Non-photorealistic illustration of surfers riding waves alongside a massive blue whale in Dana Point, stylized with vibrant colors, dynamic motion lines, and whimsical oceanic atmosphere — blending realism with fantasy art style, no photographic detail, painterly brushwork, cartoonish exaggeration of scale, serene yet adventurous mood. & \includegraphics[width=3.5cm]{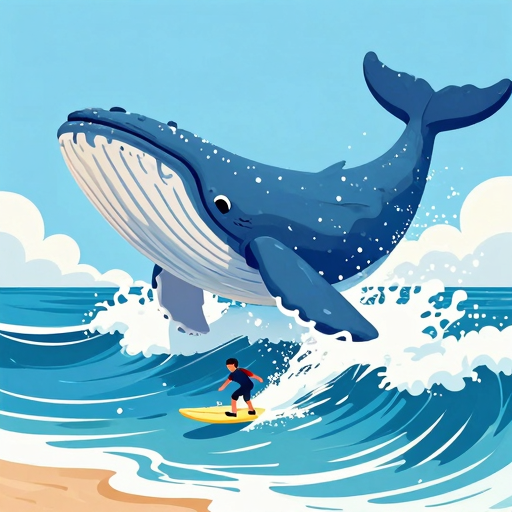} & 4 \\ 
 
1  &\footnotesize Realistic cinematic photograph of surfers riding powerful ocean waves beside a massive blue whale in Dana Point, California — dramatic natural lighting, misty coastal atmosphere, awe-inspiring scale contrast, detailed ocean textures, authentic surf gear, whale’s sleek blue-gray skin glistening, dynamic motion blur, documentary-style realism, high-resolution, 8K, environmental storytelling, capturing the surreal moment with emotional intensity and photographic accuracy. & \includegraphics[width=3.5cm]{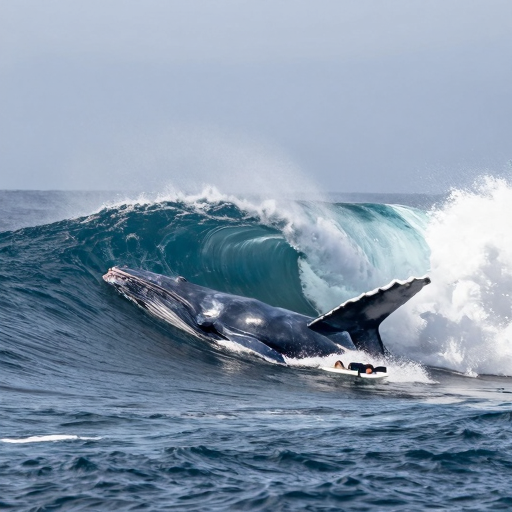} & 2 \\ 

2 &  \footnotesize Stylized cinematic illustration of surfers riding waves beside a colossal blue whale in Dana Point, California — vibrant, dreamlike atmosphere with dynamic motion, painterly brushwork, exaggerated scale for dramatic effect, whale’s glowing blue-gray skin contrasting with sunlit surfers, whimsical yet awe-inspiring mood, blending fantasy and realism, no photographic detail, high-detail ocean textures, environmental storytelling, emotional intensity, 8K resolution, dramatic coastal lighting, surreal harmony between humans and nature. & \includegraphics[width=3.5cm]{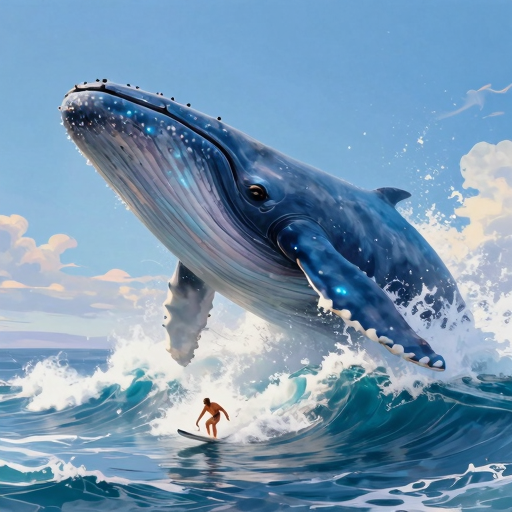} & 5  \\

\bottomrule
\end{tabular}
\end{table*}

\begin{table*}[ht]
\centering
\caption{Iterative Actor-Critic image generation for the headline \textit{``13 Popular Foods People Hated Eating 100 Years Ago.''}}
\label{tab:iterations_9682}
\setlength{\tabcolsep}{4pt}
\renewcommand{\arraystretch}{1.3}
\begin{tabular}{m{0.3cm} m{8.5cm} m{3.5cm} c  }
\toprule
\textbf{$t$} & \multicolumn{1}{>{\centering\arraybackslash}m{8.5cm}}{\textbf{Actor-produced prompt}} & {\makecell{\textbf{Generated} \\ \textbf{image}}}   & {\makecell{\textbf{Critic's} \\ \textbf{score}}} \\
\midrule

0 & \footnotesize Non-photorealistic illustration of a vintage 1920s-1930s kitchen scene with exaggerated, cartoonish expressions of disgust on people's faces as they react to bizarre or unappetizing foods like canned spinach, raw egg yolks, or processed meats — surrounded by old-fashioned kitchenware and newspaper headlines reading "13 Foods Hated 100 Years Ago," with a whimsical, retro-futuristic art style.  & \includegraphics[width=3.5cm]{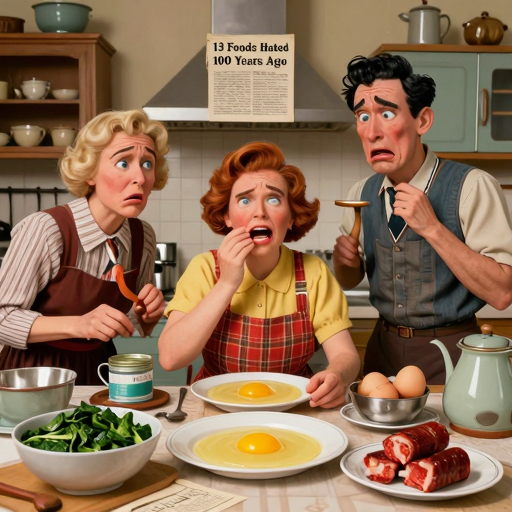} & 3 \\

1 & \footnotesize Illustrative collage of 13 bizarre, unappetizing foods from 1920s-1930s, each depicted with exaggerated cartoonish disgust reactions from people in period clothing, set against a vintage newspaper background with headlines "13 Popular Foods Hated 100 Years Ago," rendered in a whimsical, retro-futuristic comic style with bold colors and playful, exaggerated expressions.  & \includegraphics[width=3.5cm]{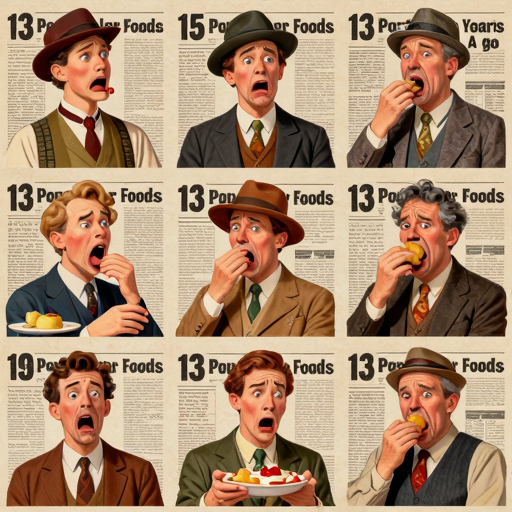} & 4 \\

2 & \footnotesize Illustrative collage featuring 13 historically unappetizing foods from the 1920s-1930s, each accompanied by exaggerated cartoonish expressions of disgust from period-dressed people, set against a vintage newspaper backdrop with the headline "13 Popular Foods Hated 100 Years Ago," rendered in a whimsical, retro-futuristic comic style with bold, saturated colors, playful exaggerated expressions, and a dynamic, eye-catching composition that emphasizes historical contrast and humor. & \includegraphics[width=3.5cm]{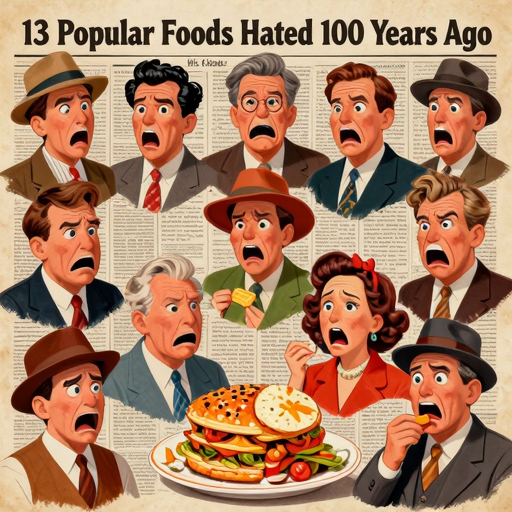} & 4  \\

3 & \footnotesize Illustrative collage of 13 historically unappetizing foods from the 1920s-1930s, each depicted with exaggerated, cartoonish expressions of disgust from people in period clothing, set against a vintage newspaper background with the bold headline "13 Popular Foods Hated 100 Years Ago," rendered in a whimsical, retro-futuristic comic style with vibrant, saturated colors, dynamic composition, playful exaggerated expressions, and a strong sense of historical humor contrasting modern palates — evoking nostalgia and amusement through visual storytelling. & \includegraphics[width=3.5cm]{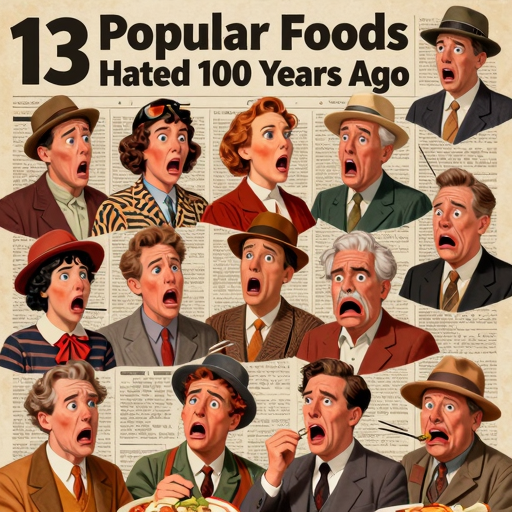} & 4  \\

4  & \footnotesize Illustrative collage of 13 historically unappetizing foods from the 1920s-1930s, each paired with exaggerated, cartoonish expressions of disgust from people in authentic period attire, set against a vintage newspaper background boldly proclaiming "13 Popular Foods Hated 100 Years Ago," rendered in a whimsical, retro-futuristic comic style with vibrant, saturated colors, dynamic composition, playful exaggerated expressions, and a strong sense of historical humor contrasting modern palates — evoking nostalgia and amusement through visual storytelling, with a touch of vintage illustration charm and bold graphic design flair.  & \includegraphics[width=3.5cm]{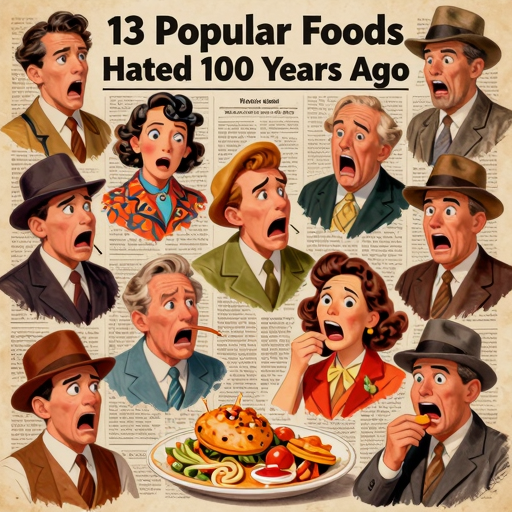} & 5\\
\bottomrule
\end{tabular}
\end{table*}

\end{document}